\documentclass{article}

\usepackage{microtype}
\usepackage{graphicx}
\usepackage{subcaption}
\usepackage{xspace}
\usepackage{booktabs} 

\usepackage{hyperref}
\usepackage{algorithm}
\usepackage{algpseudocode}

\hypersetup{%
    pdfborder = {0 0 0},
    colorlinks,
    citecolor=blue,
    linkcolor=blue,
}

\usepackage[preprint]{tmlr}

\usepackage{amsmath}
\usepackage{amssymb}
\usepackage{mathtools}
\usepackage{amsthm}
\usepackage{enumitem}

\usepackage[capitalize,noabbrev]{cleveref}

\newcommand{\dd}[2][]{\frac{\partial #1}{\partial #2}}
\newcommand{\dt}[2][]{\frac{d #1}{d #2}}
\newcommand{\de}[1]{\dt[#1]{\eta}}
\newcommand{\dl}[1]{\dt[#1]{\delta}}

\newcommand{\grad}{\operatorname{grad}}
\newcommand{\gradsq}{\operatorname{gradsq}}
\newcommand{\sgd}{\operatorname{sgd}}

\newcommand{\emam}{\operatorname{ema}_1}
\newcommand{\emav}{\operatorname{ema}_2}
\newcommand{\fu}{\operatorname{norm\_ema}}
\newcommand{\fw}{\operatorname{update}}
\newcommand{\spacev}{\phantom{m}\mathllap{v}}
\newcommand{\spacew}{\phantom{m}\mathllap{w}}

\newcommand{\eb}{\bar{\eta}}
\newcommand{\gb}{\bar{\gamma}}
\newcommand{\tb}{\bar{t}}

\newcommand{\tsum}{{\textstyle \sum}}

\newcommand{\bracket}[3]{\left#1 #3 \right#2}
\newcommand{\at}{\bracket{.}{\rvert}}
\renewcommand{\b}{\bracket{(}{)}}

\newcommand{\sqb}{\bracket{[}{]}}
\newcommand{\E}{\operatorname{E}}
\newcommand{\V}{\operatorname{Var}}

\newcommand{\mh}{\hat{m}}
\newcommand{\vh}{\hat{v}}

\newcommand{\kappap}{\kappa'}
\newcommand{\kappapp}{\kappa''}
\newcommand{\mc}{\textsc{Micro}}
\newcommand{\mb}{\textsc{micro-}batch\xspace}
\newcommand{\MB}{\textsc{mini-}batch\xspace}
\newcommand{\mbs}{\textsc{micro-}batches\xspace}
\newcommand{\MBS}{\textsc{mini-}batches\xspace}

\theoremstyle{plain}
\newtheorem{theorem}{Theorem}

\theoremstyle{definition}

\theoremstyle{remark}

\usepackage[textsize=tiny]{todonotes}

\title{Batch size invariant Adam}

\author{\name Xi Wang \email xwang3@cs.umass.edu \\
      \addr Manning College of Information and Computer Sciences\\
      University of Massachusetts Amherst
      \AND
      \name Laurence Aitchison\email laurence.aitchison@bristol.ac.uk \\
      \addr Department of Computer Science\\
      University of Bristol
    }

\begin{document}

\maketitle

\begin{abstract}
We propose a batch size invariant version of Adam, for use in large-scale, distributed settings, in which the \MB is divided into \mbs which are distributed among worker nodes.  For the $v$ term, standard Adam first computes the average over \mb gradients, then squares, while in the batch size invariant Adam proposed here, we first square the \mb gradients, then average.
Previous work (e.g. Malladi et al. 2022) used an alternative approach that involved a square-root scaling of the learning rate, but this approach requires strong assumptions to work; in particular that the gradient variance dominates the square of the expected gradient.
In contrast, the approach proposed here gives batch size invariance without this assumption.
We confirm that in practice our scheme gives batch size invariance in a much larger range of scenarios than the previous approach.
\end{abstract}

\section{Introduction}

Adam~\citep{kingma2014adam} and Adam \citep{loshchilov2017decoupled} are state-of-art optimizerss, commonly used for LLM pretraining runs \citep[e.g.][]{opt,llama}.
Given the huge investments in these pretraining runs, it is critical to understand how to tune the hyperparameters of Adam\footnote{We will use Adam to denote both Adam and AdamW for the rest of the manuscript, as our method can be applied.}.
However, as with most optimization algorithms, setting the Adam hyperparameters is difficult. 
As an example of this, we might want to tune the hyperparameters in smaller-scale settings and transfer those hyperparameters to a single, very large-scale training run.
However, to do that effectively, we need our optimization to behave in the same way for any \MB size.
This property, which we call batch size invariance, can be achieved straightforwardly in SGD, by scaling the learning rate linearly with the batch size (i.e.\ $\eta \propto B$) \citep{krizhevsky2014one,stephan2017stochastic,goyal2017accurate,smith2017bayesian,smith2017don,ma2018power,golmant2018computational,mccandlish2018empirical,zhang2019algorithmic,shallue2019measuring}.
However, for modern adaptive optimizers such as Adam \citep{kingma2014adam,loshchilov2017decoupled}, it is more difficult to achieve batch size invariance.

To understand the key barrier to obtaining truly batch size invariant Adam, we need to look at the Adam update rule,
\begin{align}
  \label{eq:adamw}
  \Delta w_{t} &= - \eta \frac{\hat{m}_t}{\sqrt{\hat{v}_t} + \epsilon}.
\end{align}
The key term here is $\hat{v}_t$, which is a debiased, exponential moving average estimate of the raw second moment of the gradient, $\E\sqb{(g')^2}$ (here, the expectation is over randomness in the choice of datapoints incorporated in the \MB, and we use $g'$ for the \MB gradient as we will use $g$ for the \mb gradient later in the manuscript).
Of course, the raw second moment is the sum of the mean-squared gradient, $\E\sqb{g'}^2$, and the variance of the gradient, $\V\sqb{g'}$.
The key issue is that the variance of the \MB gradient depends on the \MB size, $\V\sqb{g'} \propto 1/B$ \citep{granziol2022learning, malladi2022sdes,hilton2022batch}.
There is therefore an inevitable \MB size dependence in $\hat{v}_t$, which appears as a change in the effective learning rate, $\eta / (\sqrt{\hat{v}_t} + \epsilon)$.

Recent work suggested correcting for these effects by tweaking the learning rate, $\eta$ \citep{granziol2022learning,malladi2022sdes,hilton2022batch}.
In particular, they proposed a square-root scaling (i.e.\ $\eta \propto \sqrt{B}$).
However, for this square-root scaling to be valid, strong assumptions need to hold, namely that $\V\sqb{g'}$ ``dominates'' (i.e. is much bigger than) $\E\sqb{g'}^2$.
Of course, this can hardly be guaranteed to hold.
In fact, there are two settings where this assumption might break down.
First, close to initialization, where $\E\sqb{g'}$
will be large. 
Second, as the batch size gets very large (e.g. in large LLM pretraining runs), we would expect $\V\sqb{g'} \propto 1/B$ to become small.

We take a different approach to batch size invariance in Adam, by modifying the Adam updates themselves to eliminate the batch size dependence at source.
In particular, we consider a large-scale setting in which a \MB is split into a number of \mbs, which may be distributed among worker nodes.
Standard Adam computes the average gradient (across \mbs), then squares.
In contrast, we consider an alternative scheme, which first squares the \mb gradients, then averages across \mbs.
We prove that this alternative scheme is batch size invariant under much weaker assumptions. 
In particular, we do not need the gradient variance to dominate the expectation. 
We just need sufficiently small updates (e.g.\ learning rates), which is required by \citet{malladi2022sdes} anyway, and is just the Adam analogue of the critical batch size in SGD \citep{ma2018power,golmant2018computational,mccandlish2018empirical,zhang2019algorithmic,shallue2019measuring}.

%

\section{Related work}

The closest prior work is \citet{granziol2022learning}, \citet{malladi2022sdes} and \citet{hilton2022batch}, as they propose an alternative approach to obtaining batch size invariant Adam by scaling the learning rate to try to post-hoc correct for changes in the effective learning rate, $\eta / (\sqrt{\vh} + \epsilon)$, caused by changes in $\vh$.
In contrast, we propose to eliminate changes in the effective learning rate at source, by proposing a modification to Adam updates.
There are three key differences between our approach and this prior work.
First, our approach gives a linear scaling of the learning rate, $\eta \propto B$, while the prior work proposes a square-root scaling, $\eta \propto \sqrt{B}$ (this is not a contradiction: both are correct in their respective setups).
Second, our theoretical approaches are radically different.
In particular, \citet{malladi2022sdes} use stochastic differential equations.
In particular, they take the choice of datapoints in the \MB to be random, and they consider the updates to be random variables.
In contrast, in our actual proof (Appendix~\ref{app:proof_1}), we consider the choice of datapoints in the \MBS as fixed, and thus the updates become deterministic.
Instead, we ask: if we give the same set of datapoints to two optimizers, in what settings are the resulting weight updates exactly equivalent?
Third, our approach gives batch size invariance under much weaker assumptions. 
In particular, we do not need the gradient variance to dominate the expectation, we only need the updates to be sufficiently small (which is required by \citet{malladi2022sdes} anyway).

It turns out that this requirement for the updates to be sufficiently small is also encountered in SGD, where it is understood as a critical batch size \citep{ma2018power}.
Notably, this work assumes takes $\eta \propto B$ so a critical batch size is intimately related to a critical learning rate.
\citet{shallue2019measuring} introduced the term ``perfect'' scaling for the region in which training is batch size invariant and studied the effect of e.g.\ architectures on the critical batch size.
The dependence of the critical batch size on dataset \citep{golmant2018computational}, gradient noise scale \citep{mccandlish2018empirical} and curvature \citet{zhang2019algorithmic}.
We observe a similar threshold for batch size invariant Adam (Fig.~\ref{fig:resnet_ln}), though in our plots it is most straightforwardly interpreted as a critical learning rate.

The notion of computing the squared gradients for \mbs was also proposed in \citet{zhang2023adam}, albeit in a very different context.
In particular, they were trying to do Adam on large \MBS in memory constrained, single-node settings.
They therefore considered splitting the \MB into \mbs, and aggregating gradients across the \mb.
They found that they were able to reduce memory consumption by directly aggregating squared \mb gradients, rather than by aggregating gradients and squaring at the end.
However, given common results that smaller batches tend to work better \citep[e.g.][]{keskar2016large}, it is not clear why you would do aggregate across \mbs on a single node: why not just apply Adam updates to the \mbs?
This is confirmed by our results, which show that as updates get too large, and invariance breaks down and smaller batches converge faster (Fig.~\ref{fig:resnet_ln}).

Importantly, \citet{zhang2023adam} is entirely focused on reducing memory consumption: they do not consider batch size invariance, and such, their proposed algorithm is not batch size invariant.
In particular, they consider reducing memory consumption by taking a fixed-size \MB and splitting it into an increasingly large number of \mbs.
Of course, as we have more \mbs, each \mb gets smaller, so the variance of the \mb gradient estimate gets larger, so $\vh$ will increase and the effective learning rate will fall.
Our contribution is therefore entirely orthogonal: establishing that a particular variant of Adam with a particular choice of hyperparameter scaling is invariant to batch size for small enough updates.

\begin{figure}[t]

\small
\centering
\begin{tabular}{ll}
\hline
\textbf{Symbol} & \textbf{Description} \\
\hline
$\kappa$           & Number of \mbs in a \MB. \\
$w_t$           & Parameter values after consuming $t$ \mb. \\
$g_t(w)$            & Gradient for the $t$th \mb evaluated at $w$. \\
$\eta$       & Learning rate. \\
$(\gamma_1, \gamma_2)$  & EMA parameters (related to $\beta_1$ and $\beta_2$, Eq.~\ref{eq:gammas}).\\
$\lambda$  & Weight decay coefficient.\\
\hline
\end{tabular}

\begin{algorithm}[H]
   \caption{\mc~Adam (i.e.\ standard Adam applied to \mbs)}
   \label{alg:micro_adamw}
\begin{algorithmic}
   \While{not converged}
     \State $m_{t} = (1-\gamma_1) m_{t-1} + \gamma_1 g_{t}(w_{t-1})$
     \State $\spacev_{t} = (1-\gamma_2) \spacev_{t-1} + \textcolor{red}{\gamma_2 g_{t}^2(w_{t-1})}$
     \State $\mh_{t} = \frac{m_t}{1-\beta_1(\gamma_1)^{t}}$
     \State $\hat{\spacev}_{t} = \frac{v_t}{1-\beta_2(\gamma_2)^{t}}$
     \State $\spacew_t = (1- \eta \lambda) w_{t-1} - \eta \frac{\hat{m}_t}{\sqrt{\hat{v}_t} + \epsilon}$
   \EndWhile
\end{algorithmic}
\end{algorithm}

\vspace{-1.8 \baselineskip}

\begin{algorithm}[H]
   \caption{Standard Adam applied to \MBS.}
   \label{alg:standard_adamw}
\begin{algorithmic}
   \While{not converged}
     \State $m_{t} = (1-\gamma_1') m_{t-\kappa} + \gamma_1' \tfrac{1}{\kappa} \tsum_{k=1}^\kappa g_{t-k+\kappa}(w_{t-\kappa})$
     \State $\spacev_{t} = (1-\gamma_2') \spacev_{t-\kappa} + \gamma_2' \textcolor{red}{(\tfrac{1}{\kappa} \tsum_{k=1}^\kappa g_{t-k+\kappa}(w_{t-\kappa}))^2}$
     \State $\mh_{t} = \frac{m_t'}{1-\beta_1(\gamma_1')^{t/\kappa}}$
     \State $\hat{\spacev}_{t} = \frac{v_t'}{1-\beta_2(\gamma_2')^{t/\kappa}}$
     \State $\spacew_t = (1- \eta' \lambda) w_{t-\kappa} - \eta' \frac{\hat{m}_t}{\sqrt{\hat{v}_t} + \epsilon}$
   \EndWhile
\end{algorithmic}
\end{algorithm}

\vspace{-1.8 \baselineskip}

\begin{algorithm}[H]
   \caption{Batch size invariant Adam applied to \MBS.}
   \label{alg:invariant_adamw}
\begin{algorithmic}
   \While{not converged}
     \State $m_{t} = (1-\gamma_1') m_{t-\kappa} + \gamma_1' \tfrac{1}{\kappa} \tsum_{k=1}^\kappa g_{t-k+\kappa}(w_{t-\kappa})$
     \State $\spacev_{t} = (1-\gamma_2') \spacev_{t-\kappa} + \gamma_2' \textcolor{red}{\tfrac{1}{\kappa} \tsum_{k=1}^\kappa g_{t-k+\kappa}^2(w_{t-\kappa})}$
     \State $\mh_{t} = \frac{m_t'}{1-\beta_1(\gamma_1')^{t/\kappa}}$
     \State $\hat{\spacev}_{t} = \frac{v_t'}{1-\beta_2(\gamma_2')^{t/\kappa}}$
     \State $\spacew_t = (1- \eta' \lambda) w_{t-\kappa} - \eta' \frac{\hat{m}_t}{\sqrt{\hat{v}_t} + \epsilon}$
   \EndWhile
\end{algorithmic}
\end{algorithm}

\vspace{-2 \baselineskip}

\end{figure}

That said, the key insight in \cite{zhang2023adam} is that you can avoid the memory cost of explicitly accumulating the average gradient by immediately incorporating the \mb gradients into $m_t$ and the \mb square gradients into $v_t$ (which is possible in Alg.~\eqref{alg:invariant_adamw} as the $m_t$ and $v_t$ updates are just sums over \mbs).
While \citet{zhang2023adam} did not consider the multi-node setting, the key insight holds true in multi-node settings for the server that aggregates gradient updates.
In particular, this server again does not need to retain a gradient accumulator, but can just accumulate directly into $m_t$ and $v_t$.  
This raises the prospect of a potential memory saving of 25\% at the server (in standard Adam, we need to accumulate four quantities for each parameter, $g'_t$, $m_t$, $v_t$ and $w_t$, while if you use squared \mb gradients, you only need $m_t$, $v_t$ and $w_t$).
That raises the tantalising prospect that the batch size invariant Adam considered here is also more efficient in practice in large-scale distributed settings.

We frame our proposal in terms of \mbs computed on separate workers, as that makes it easy to compute the squared gradient for the \mb on a worker.
However, our proposal can also be used in the single-worker setting.
For instance, it would be possible to treat each datapoint as a \mb, and compute the sum of squared gradients for individual datapoints using techniques from 
\citet{dangel2020backpack} (which critically avoid needing to materialise the full tensor of individual datapoint gradients, which has an impractically large size of $B \times$  number of parameters).

\section{Methods}

To set the context, we introduce the concept of \mb. In particular, we assume a \MB of $B$ samples can be equally split into $\kappa$ \mbs, and then we aggregate $\kappa$ sets of gradient computed at \emph{different} microbaches under the \emph{same} parameter values for the actual parameter updates.
The idea of \mb is widely adopted in the training of deep networks, e.g. {\mb}ing facilitates larger \MB sizes under constrained memory by enabling gradient accumulation and supports data parallelism through distributing \mbs across workers for parallel gradient computation. 
However, our focus diverges from these applications; we leverage {\mb}ing to explore how optimization algorithms are influenced by \MB size, i.e.\ value of $\kappa$. Notice that, we assume a \emph{fixed} \mb size through the analysis and we will use batch size to denote the \emph{\mb} size in later paragraphs.

We begin by writing down standard Adam updates applied to individual \mbs (denoted $t$) (Alg.~\ref{alg:micro_adamw}).
Here, $g_t(w)$ is the gradient of the objective for the $t$th \mb.
Note that we have written Adam here in a slightly unusual parameterization, using,
\begin{align}
  \label{eq:gammas}
  \beta_1(\gamma_1) &= 1-\gamma_1 &
  \beta_2(\gamma_2) &= 1-\gamma_2
\end{align}
as the scalings we derive later on are much more simply expressed in terms of $\gamma_1$ and $\gamma_2$ than in terms of $\beta_1$ and $\beta_2$.
Note that we can recover Adam from AdamW by setting $\lambda=0$, and incorporating any AdamW weight decay into the objective.

Standard Adam updates (Alg.~\ref{alg:standard_adamw}) are performed on a \MB, formed by merging $\kappa$ \mbs.
Importantly, to write down these merged, updates we continue to take $t$ to be \textit{\mbs}, rather than steps or \MBS.
Of course, each update of standard Adam consumes a single \MB consisting of $\kappa$ \mbs. 
Thus, if we start at $w_{t-\kappa}$ then a single step of standard Adam will consume $\kappa$ \mbs, taking us to $w_{t}$. 
Another step consumes another $\kappa$ \mbs, taking us to $w_{t+\kappa}$.
Thus, while $t$ indexes the number of \mbs, $t/\kappa$ indexes optimizer steps.
This unusual convention has the key benefit that the time indices in standard Adam remains aligned with the time indices in \mc~Adam (Alg.~\ref{alg:micro_adamw}), which will be important as we will be assessing equivalence between \mc~Adam and standard Adam or batch size invariant Adam.
However, this notation does have the disadvantage that it modifies slightly the form of the debiasing equations (see the powers of $\text{steps} = t/\kappa$ in the expressions for $\mh$ and $\vh$ in Alg.~\eqref{alg:standard_adamw} and Alg.~\eqref{alg:invariant_adamw}).

\paragraph{Batch size dependence in standard Adam.} A standard Adam update (Alg.~\ref{alg:standard_adamw}) applies to a \textit{\MB} formed by merging $\kappa$ \mbs.
However, the arguments given in the introduction indicate that standard Adam is not batch size invariant.
By ``batch size invariant'', we mean an algorithm for which, in some sensible limit, a single update on a \MB formed of $\kappa$ \mbs is equivalent to $\kappa$ update steps of \mc~Adam on the underlying \mbs (Alg.~\ref{alg:micro_adamw}).
To see that standard Adam (Alg.~\ref{alg:standard_adamw}) is not batch size invariant, we consider the $\vh_t$ terms in the various algorithms. 
For \mc~Adam (Alg.~\ref{alg:micro_adamw}), $\vh_t$ is an exponential moving average of the raw second moment of the \textit{\mb} gradients,
\begin{align}
  \label{eq:microbatch_sq_grad}
  \E\sqb{g_t^2(w_{t-1})} &= \E\sqb{g}^2 + \V\sqb{g}.
\end{align}
where $\E\sqb{g}$ and $\V\sqb{g}$ is the mean and variance of \mb gradients.
In contrast, for standard Adam (Alg.~\ref{alg:standard_adamw}), $\vh_t$ is an exponential moving average of the raw second moment of the \textit{minibatch} gradients
\begin{align}
\label{eq:minibatch_sq_grad}
\E\sqb{(\tfrac{1}{\kappa} \tsum_{k=1}^\kappa g_{t-k+\kappa}(w_{t-\kappa}))^2} &= \E\sqb{g}^2 + \tfrac{1}{\kappa} \V\sqb{g}.
\end{align}
Critically, as we have more \mbs, $\kappa$, in the \MB, we get a better estimate of the gradient and hence the variance term falls.  
And as the effective learning rate, $\eta / (\sqrt{\vh_t} + \epsilon)$ depends on $\vh_t$, we can see that the effective learning rate depends unavoidably on the batch size.  

\paragraph{Batch size invariant Adam.} We therefore consider a batch size invariant version of Adam (Alg.~\ref{alg:invariant_adamw}) in which the red term is the average over squared \mb gradients (as opposed to standard Adam, where the red term is the square of average \mb gradients).
Thus, in batch size invariant Adam (Alg.~\ref{alg:invariant_adamw}), just like in \mc~Adam (Alg.~\ref{alg:micro_adamw}), $\vh$ estimates the expected raw second moment of the fixed-size \mb gradients,
\begin{align}
\E\sqb{\tfrac{1}{\kappa} \tsum_{k=1}^\kappa g_{t-k+\kappa}^2(w_{t-\kappa})} &= \E\sqb{g}^2 + \V\sqb{g}
\end{align}
This expectation thus does not depend on $\kappa$, avoiding a key source of batch size dependence.

\section{Theorem statements}

While we have clearly eliminated one source of batch size dependence, we have yet to prove that Alg.~\eqref{alg:invariant_adamw} really is batch size invariant in any formal sense.
We begin with a theorem that states that with an appropriate choice of hyperparameters, $\kappa$ steps of \mc~Adam on \mbs of size $M$ gives equivalent optimizer-state-updates (i.e.\ updates for $m_t$, $v_t$ and $g_t$) to a single step of batch size invariant Adam with the same datapoints grouped into a single \MB of size $B=\kappa M$ (Theorem~\ref{thm:micro_bi_equiv}).
This implies that with an appropriate choice of hyperparameters, batch size invariant Adam with different \MB sizes gives equivalent optimizer-state-updates after consuming the same set of data points (Theorem~\ref{thm:batch size_invariance}).
For the proofs, see Appendices~\ref{app:grads}--\ref{app:proof_2}.


To understand the formal statement of the theorem, we need to understand how to formally obtain small $\eta$, $\gamma_1$ and $\gamma_2$.
We do this by setting,
\begin{subequations}
\begin{align}
  \eta &= \delta \eb_0\\
  \gamma_1 &= \delta \gb_1\\
  \gamma_2 &= \delta \gb_2.
\end{align}
\end{subequations}
and taking $\delta \rightarrow 0$.
Of course, as $\delta \rightarrow 0$, all state variables ($w_t$, $m_t$ and $v_t$) stop changing so the updates from the two optimizers are trivially equivalent.
To avoid this trivial equivalence, we actually consider the equivalence of weight changes, divided by $\delta$ (see below).

\begin{theorem}
\label{thm:micro_bi_equiv}
Consider two optimizers: \mc~Adam (Alg.~\ref{alg:micro_adamw}; i.e.\ standard Adam applied to \mbs) with hyperparameters $\eta$, $\gamma_1$ and $\gamma_2$, and batch size invariant Adam (Alg.~\ref{alg:invariant_adamw}) with hyperparameters
\begin{subequations}
\label{eq:invariant_adamw_hyper}
\begin{align}
  \eta' &= \kappa \eta,\\
  \gamma_1' &= \kappa \gamma_1,\\ 
  \gamma_2' &= \kappa \gamma_2.
\end{align}
\end{subequations}
applied to \MBS composed of $\kappa$ \mbs.
We start both optimizers at time $t-\kappa$ at the same initial state, $w_{t-\kappa}$, $m_{t-\kappa}$ and $v_{t-\kappa}$.
We take $w_t$ to be the result of applying $\kappa$ steps of standard Adam to $\kappa$ \mbs, and $w'_t$ to be the result of applying a single step of batch size invariant Adam to a \MB (consisting of the same $\kappa$ \mbs merged together).
Then,
\begin{subequations}
\label{eq:thm}
\begin{align}
  \lim_{\delta \rightarrow 0} \frac{m_t - m_{t-\kappa}}{\delta} &= \lim_{\delta \rightarrow 0} \frac{m'_t - m_{t-\kappa}}{\delta}\\
  \lim_{\delta \rightarrow 0} \frac{v_t - v_{t-\kappa}}{\delta} &= \lim_{\delta \rightarrow 0} \frac{v'_t - v_{t-\kappa}}{\delta}\\
  \lim_{\delta \rightarrow 0} \frac{w_t - w_{t-\kappa}}{\delta} &= \lim_{\delta \rightarrow 0} \frac{w'_t - w_{t-\kappa}}{\delta}.
\end{align}
\end{subequations}
i.e.\ the state updates for the merged and unmerged optimizers are equivalent for sufficiently small $\eta$, $\gamma_1$ and $\gamma_2$.
\end{theorem}

\begin{theorem}
\label{thm:batch size_invariance}
Consider batch size invariant Adam under two \MB sizes $B' = \kappap M$ and $B'' = \kappapp M$, where $M$ denotes the size of the \mb, with hyperparameters
\begin{subequations}
\begin{align}
  \eta' &= \kappap \eta_0, &\eta'' &= \kappapp \eta_0\\
  \gamma_1' &= \kappap \gamma_1, &\gamma_1'' &= \kappapp \gamma_1 \\ 
  \gamma_2' &= \kappap \gamma_2, &\gamma_2'' &= \kappapp \gamma_2
\end{align}    
\end{subequations}
Consider an integer number of \mbs, $K$, which can be divided by both $\kappap$ and $\kappapp$.
We start both optimizers at time $t - K$ with state $m_{t-K},v_{t-K},w_{t-K}$.
We take $K / \kappap$ update steps with the first optimizer, which results in an optimizer state of $m_t', v_t', w_t'$. 
We also take $K / \kappapp$ update steps with the second optimizer, which results in an optimizer state of $m_t'', v_t'', w_t''$.
Then,
\begin{subequations}
\begin{align}
  \lim_{\delta \rightarrow 0} \frac{m_t'- m_{t-K}}{\delta} &= \lim_{\delta \rightarrow 0} \frac{m_t'' - m_{t-K}}{\delta}\\
  \lim_{\delta \rightarrow 0} \frac{v_t' - v_{t-K}}{\delta} &= \lim_{\delta \rightarrow 0} \frac{v_t'' - v_{t-K}}{\delta}\\
  \lim_{\delta \rightarrow 0} \frac{w_t' - w_{t-K}}{\delta} &= \lim_{\delta \rightarrow 0} \frac{w_t'' - w_{t-K}}{\delta}.
\end{align}
\end{subequations}
i.e.\ the state updates are equivalent for sufficiently small $\eta$, $\gamma_1$ and $\gamma_2$.
\end{theorem} 

\section{Experiments}

\begin{figure*}[p]
    \centering
    \includegraphics[width=0.9\textwidth]{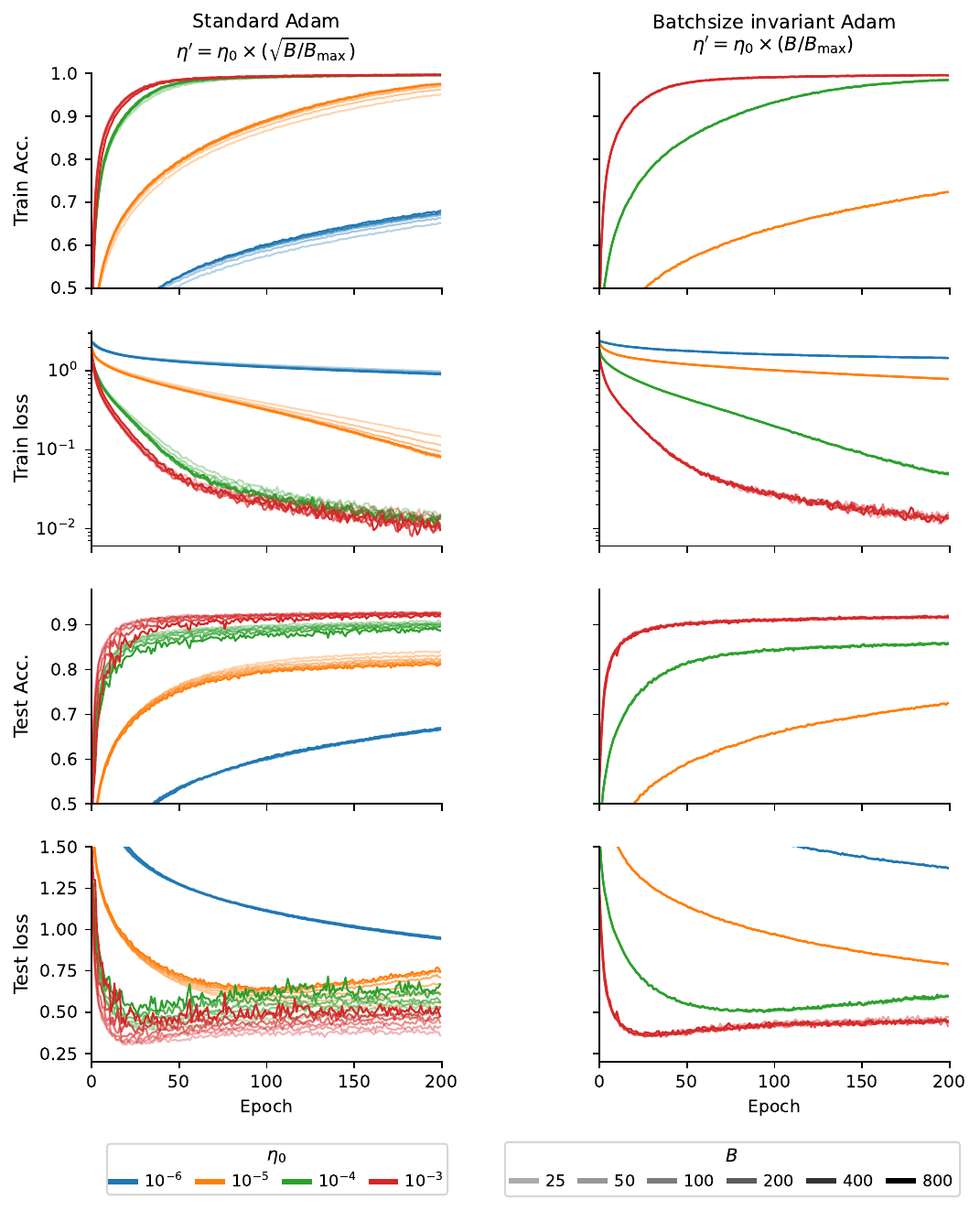}
     \caption{Comparing the behavior of our proposed batch size invariant Adam (right column), with $\eta \propto B$ against standard Adam (left column), with $\eta \propto \sqrt{B}$
     \citep{granziol2022learning,malladi2022sdes,hilton2022batch}. The model was a ResNet-18 trained on CIFAR-10 over 200 epochs, under batch sizes (opacity) ranging from $B=25$ to $B=800$, with $B_\mathrm{max}=800$. Each color represents a different base learning rate $\eta_0$. Note that batch size invariant Adam (right) gives almost perfect batch size invariance (in that the lines are all on top of each other) up until $\eta=10^{-3} \times (B/B_\text{max})$.  In contrast, with standard Adam (bottom), you get discrepancies even with the smallest learning rate, i.e. $\eta_0 = 10^{-6}$.
  }
    \label{fig:resnet}
\end{figure*}

\begin{figure*}[p]
    \centering
    \includegraphics[width=0.9\textwidth]{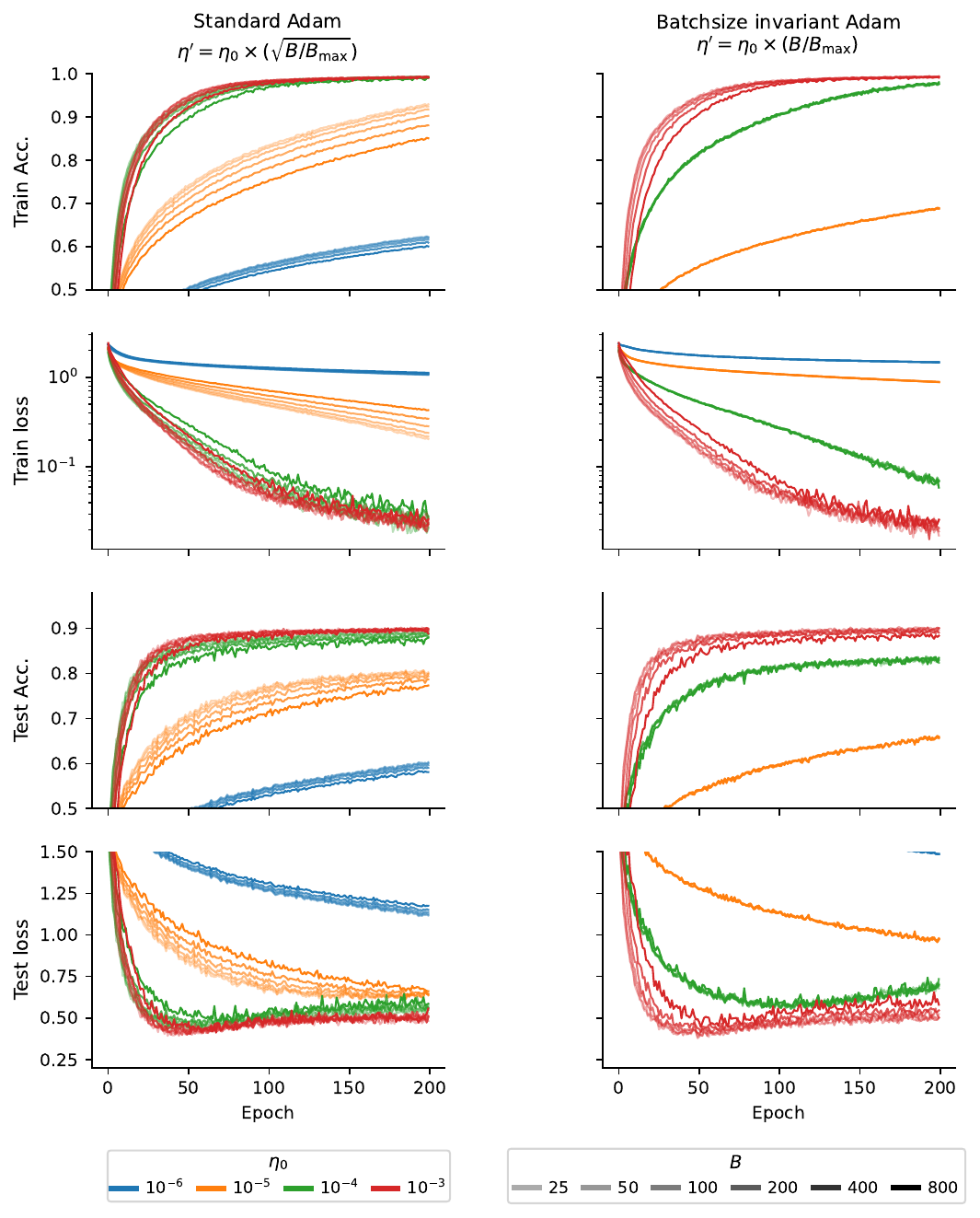}
    \caption{As Fig.~\ref{fig:resnet}, but with layernorm rather than batchnorm. In particular, we compare the behavior of our proposed batch size invariant Adam, with $\eta \propto B$ (right) against standard Adam, with $\eta \propto \sqrt{B}$ (left), both with $B_{\rm max} = 800$. Similar to the batchnorm results, the batch size invariant Adam lines (right) almost exactly line up, until $\eta=10^{-3} \times (B/B_\text{max})$. Whereas standard Adam (left) shows discrepancies between lines even under the smallest learning rate considered ($\eta_0 = 10^{-6}$).}
    \label{fig:resnet_ln}
\end{figure*}

\begin{figure*}[!ht]
    \centering
    \includegraphics[width=0.9\textwidth]{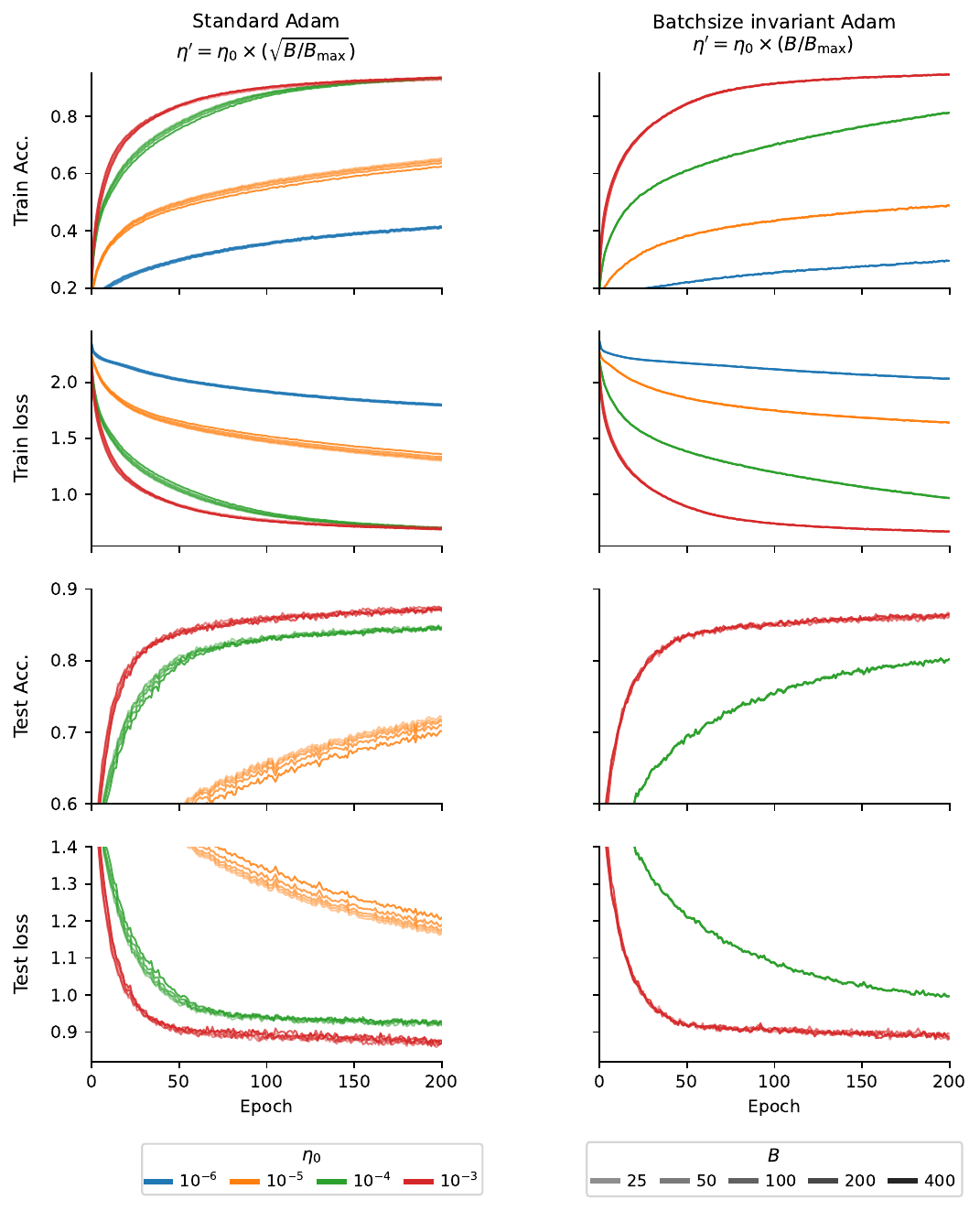}
    \caption{As Fig.~\ref{fig:resnet}, but use ViT rather than ResNet-18. Again, we compare the behavior of our proposed batch size invariant Adam, with $\eta \propto B$ (right) against standard Adam, with $\eta \propto \sqrt{B}$ (left). A $B_\text{max}$ of 400 is used. Standard Adam shows aligned trajectories for the smallest and largest base learning rate but shows significant discrepancy at $\eta_0 = 10^{-5}$ and in the early stage of $\eta_0=10^{-4}$. This is expected, as obtaining batch size invariance with standard Adam under the square-root scaling requires the gradient variance dominating the gradient mean, which may not hold when the parameters are near initialization, where the gradient may have a large magnitude. Regardless, our proposed batchsize invariant Adam shows consistency across all learning rates considered at all stages in the  optimization.}
    \label{fig:vit}
\end{figure*}

In this section, we empirically confirmed that batch size invariant Adam with sufficiently small learning rates indeed gave almost exactly equivalent optimization results under different batch sizes. We additionally compared its optimization trajectories against standard Adam under the square root scaling rule, where we observed a higher level of inœconsistency across different batch sizes.

\paragraph{Task description} We trained ResNet-18~\citep{he2016deep} and vision transformer~\citep[ViT]{dosovitskiy2020image} on CIFAR-10~\citep{krizhevsky2009learning}. For ResNet-18, we used standard cross-entropy loss with random cropping and flipping as data augmentation. For ViT\footnote{Implementation based on \url{https://github.com/kuangliu/pytorch-cifar/}}, we used a patch size of 8 and we added QK-layernorm~\citep{dehghani2023scaling, wortsman2023small} for more stabilized training under larger learning rates. We additionally incorporated label smoothing loss and automatic data augmentation~\citep{cubuk2018autoaugment} to match up the test performance with the numbers reported in other literature, e.g. ~\citet{schaipp2023momo}.

Unfortunately, batchnorm itself can introduce batch size dependent effects (which arise because larger batches imply more accurate estimates of the feature means and variances).
We therefore considered the ResNet-18 with batchnorm (Fig.~\ref{fig:resnet}) and replacing batchnorm with layernorm (Fig.~\ref{fig:resnet_ln}).
However, in our experiments, the dominant batch size dependent effects appeared to arise from Adam (Fig.~\ref{fig:resnet}), as we were able to dramatically reduce batch size dependence using our batch size invariant Adam.

\paragraph{Optimization setting} We considered two optimization methods: Batch size invariant Adam with a \mb size of 25, with actual learning rate $\eta'$ scaled linearly with the batch size $B$:
\begin{align}
    \eta' &= \eta_0 B / B_{\rm max},
\end{align}
and standard Adam, with $\eta'$ proportion to $\sqrt{B}$:
\begin{align}
    \eta' &= \eta_0 \sqrt{B / B_{\rm max}}.
\end{align}
where $\eta_0$ is the base learning rate and $B_{\rm max}$ is the largest batch size used for a particular task.
For both methods, we set the running average parameters \citep{hilton2022batch} as
\begin{align}
  \gamma_1' &= 0.1 \times \tfrac{B}{B_\text{max}} &
  \gamma_2' &= 0.001 \times \tfrac{B}{B_\text{max}}.
\end{align}
For both tasks and optimizers, we experimented with $\eta_0$ in range $\{10^{-6},10^{-5},10^{-4},10^{-3}\}$. We tested $B$ between $25$ to $800$ for ResNet and $B$ between $25$ to $400$ for ViT, where the largest batch sizes (i.e. $B_{\rm max}$) are chosen as the largest value that can fit into a single 2080ti GPU.
Lastly, we used constant learning rates without scheduling and no weight decay during optimization.

\paragraph{Results} The results are presented in Fig.~\ref{fig:resnet} (ResNet with batchnorm), Fig.~\ref{fig:resnet_ln} (ResNet with layernorm) and Fig.~\ref{fig:vit} (ViT). All figures show the model's performance metrics against the training epoch for different learning rates (color) and batch sizes (opacity) averaged over three random trials.

On ResNet, under our proposed batch size invariant Adam (Fig.~\ref{fig:resnet},~\ref{fig:resnet_ln} right), the trajectories under different batch sizes almost exactly overlap for all learning rates with batchnorm (Fig.~\ref{fig:resnet} right), and for all except the largest learning rate with layernorm (Fig.~\ref{fig:resnet_ln} right) (note that we do expect batch size invariance to break down at some point as the learning rate increases).
In contrast, the standard Adam with square-root scaling (Fig.~\ref{fig:resnet} left) reveals more pronounced differences across batch sizes across a far wider range of learning rates.

Next, we considered ViT.
Similar to ResNet results, our proposed batch size invariant Adam (Fig.~\ref{fig:vit} right) shows aligned trajectories across different batch sizes for all learning rates, whereas standard Adam (Fig.~\ref{fig:vit} left) shows more pronounced differences, which are especially evident for $\eta_0=10^{-5},10^{-4}$.

One thing that may be evident from the figures is that the rate of convergence for different settings of $\eta_0$ is not comparable between standard Adam and batch size-invariant Adam (i.e.\ comparing lines of the same color on the left and right sides of Figs.~\ref{fig:resnet}--\ref{fig:vit}).
This is expected, even for $B = B_\text{max}$, so $\eta'=\eta_0$ for both standard Adam and batch size invariant Adam.
The reason is that the root-mean-squared gradient denominators in the update rules are different.
Standard Adam takes the average over the whole \MB before computing the root-mean-squared average, which reduces the impact of variance in the gradient (Eq.~\eqref{eq:minibatch_sq_grad}).
In contrast, batch size invariant Adam takes the root-mean-square average for \mbs (Eq.~\eqref{eq:microbatch_sq_grad}), which retains contribution from variance in the gradient, even as the batch size grows.

\section{Conclusions}
We developed a batch size invariant Adam for large-scale distributed settings, which computes the expectation of the squared \textit{microbatch} gradients, rather than the squared \textit{minibatch} gradients used by standard Adam.
We proved batch size invariance under mild conditions (just sufficiently small updates), without needing to go to stochastic differential equations.
Furthermore, insights from \citep{zhang2023adam} suggest that this approach may even be more efficient than standard Adam in practical, large-scale settings.


\bibliographystyle{icml2024}
\bibliography{example_paper}

\begin{thebibliography}{27}
\providecommand{\natexlab}[1]{#1}
\providecommand{\url}[1]{\texttt{#1}}
\expandafter\ifx\csname urlstyle\endcsname\relax
  \providecommand{\doi}[1]{doi: #1}\else
  \providecommand{\doi}{doi: \begingroup \urlstyle{rm}\Url}\fi

\bibitem[Cubuk et~al.(2018)Cubuk, Zoph, Mane, Vasudevan, and Le]{cubuk2018autoaugment}
Cubuk, E.~D., Zoph, B., Mane, D., Vasudevan, V., and Le, Q.~V.
\newblock Autoaugment: Learning augmentation policies from data.
\newblock \emph{arXiv preprint arXiv:1805.09501}, 2018.

\bibitem[Dangel et~al.(2020)Dangel, Kunstner, and Hennig]{dangel2020backpack}
Dangel, F., Kunstner, F., and Hennig, P.
\newblock Back{PACK}: Packing more into backprop.
\newblock In \emph{International Conference on Learning Representations}, 2020.

\bibitem[Dehghani et~al.(2023)Dehghani, Djolonga, Mustafa, Padlewski, Heek, Gilmer, Steiner, Caron, Geirhos, Alabdulmohsin, et~al.]{dehghani2023scaling}
Dehghani, M., Djolonga, J., Mustafa, B., Padlewski, P., Heek, J., Gilmer, J., Steiner, A.~P., Caron, M., Geirhos, R., Alabdulmohsin, I., et~al.
\newblock Scaling vision transformers to 22 billion parameters.
\newblock In \emph{International Conference on Machine Learning}, pp.\  7480--7512. PMLR, 2023.

\bibitem[Dosovitskiy et~al.(2020)Dosovitskiy, Beyer, Kolesnikov, Weissenborn, Zhai, Unterthiner, Dehghani, Minderer, Heigold, Gelly, et~al.]{dosovitskiy2020image}
Dosovitskiy, A., Beyer, L., Kolesnikov, A., Weissenborn, D., Zhai, X., Unterthiner, T., Dehghani, M., Minderer, M., Heigold, G., Gelly, S., et~al.
\newblock An image is worth 16x16 words: Transformers for image recognition at scale.
\newblock In \emph{International Conference on Learning Representations}, 2020.

\bibitem[Golmant et~al.(2018)Golmant, Vemuri, Yao, Feinberg, Gholami, Rothauge, Mahoney, and Gonzalez]{golmant2018computational}
Golmant, N., Vemuri, N., Yao, Z., Feinberg, V., Gholami, A., Rothauge, K., Mahoney, M.~W., and Gonzalez, J.
\newblock On the computational inefficiency of large batch sizes for stochastic gradient descent.
\newblock \emph{arXiv preprint arXiv:1811.12941}, 2018.

\bibitem[Goyal et~al.(2017)Goyal, Doll{\'a}r, Girshick, Noordhuis, Wesolowski, Kyrola, Tulloch, Jia, and He]{goyal2017accurate}
Goyal, P., Doll{\'a}r, P., Girshick, R., Noordhuis, P., Wesolowski, L., Kyrola, A., Tulloch, A., Jia, Y., and He, K.
\newblock Accurate, large minibatch sgd: Training imagenet in 1 hour.
\newblock \emph{arXiv preprint arXiv:1706.02677}, 2017.

\bibitem[Granziol et~al.(2022)Granziol, Zohren, and Roberts]{granziol2022learning}
Granziol, D., Zohren, S., and Roberts, S.
\newblock Learning rates as a function of batch size: A random matrix theory approach to neural network training.
\newblock \emph{The Journal of Machine Learning Research}, 23\penalty0 (1):\penalty0 7795--7859, 2022.

\bibitem[He et~al.(2016)He, Zhang, Ren, and Sun]{he2016deep}
He, K., Zhang, X., Ren, S., and Sun, J.
\newblock Deep residual learning for image recognition.
\newblock In \emph{Proceedings of the IEEE conference on computer vision and pattern recognition}, pp.\  770--778, 2016.

\bibitem[Hilton et~al.(2022)Hilton, Cobbe, and Schulman]{hilton2022batch}
Hilton, J., Cobbe, K., and Schulman, J.
\newblock Batch size-invariance for policy optimization.
\newblock \emph{Advances in Neural Information Processing Systems}, 35:\penalty0 17086--17098, 2022.

\bibitem[Keskar et~al.(2016)Keskar, Mudigere, Nocedal, Smelyanskiy, and Tang]{keskar2016large}
Keskar, N.~S., Mudigere, D., Nocedal, J., Smelyanskiy, M., and Tang, P. T.~P.
\newblock On large-batch training for deep learning: Generalization gap and sharp minima.
\newblock \emph{arXiv preprint arXiv:1609.04836}, 2016.

\bibitem[Kingma \& Ba(2014)Kingma and Ba]{kingma2014adam}
Kingma, D.~P. and Ba, J.
\newblock Adam: A method for stochastic optimization.
\newblock \emph{arXiv preprint arXiv:1412.6980}, 2014.

\bibitem[Krizhevsky(2014)]{krizhevsky2014one}
Krizhevsky, A.
\newblock One weird trick for parallelizing convolutional neural networks.
\newblock \emph{arXiv preprint arXiv:1404.5997}, 2014.

\bibitem[Krizhevsky et~al.(2009)Krizhevsky, Hinton, et~al.]{krizhevsky2009learning}
Krizhevsky, A., Hinton, G., et~al.
\newblock Learning multiple layers of features from tiny images.
\newblock 2009.

\bibitem[Loshchilov \& Hutter(2017)Loshchilov and Hutter]{loshchilov2017decoupled}
Loshchilov, I. and Hutter, F.
\newblock Decoupled weight decay regularization.
\newblock \emph{arXiv preprint arXiv:1711.05101}, 2017.

\bibitem[Ma et~al.(2018)Ma, Bassily, and Belkin]{ma2018power}
Ma, S., Bassily, R., and Belkin, M.
\newblock The power of interpolation: Understanding the effectiveness of sgd in modern over-parametrized learning.
\newblock In \emph{International Conference on Machine Learning}, pp.\  3325--3334. PMLR, 2018.

\bibitem[Malladi et~al.(2022)Malladi, Lyu, Panigrahi, and Arora]{malladi2022sdes}
Malladi, S., Lyu, K., Panigrahi, A., and Arora, S.
\newblock On the sdes and scaling rules for adaptive gradient algorithms.
\newblock \emph{Advances in Neural Information Processing Systems}, 35:\penalty0 7697--7711, 2022.

\bibitem[McCandlish et~al.(2018)McCandlish, Kaplan, Amodei, and Team]{mccandlish2018empirical}
McCandlish, S., Kaplan, J., Amodei, D., and Team, O.~D.
\newblock An empirical model of large-batch training.
\newblock \emph{arXiv preprint arXiv:1812.06162}, 2018.

\bibitem[Schaipp et~al.(2023)Schaipp, Ohana, Eickenberg, Defazio, and Gower]{schaipp2023momo}
Schaipp, F., Ohana, R., Eickenberg, M., Defazio, A., and Gower, R.~M.
\newblock Momo: Momentum models for adaptive learning rates.
\newblock \emph{arXiv preprint arXiv:2305.07583}, 2023.

\bibitem[Shallue et~al.(2019)Shallue, Lee, Antognini, Sohl-Dickstein, Frostig, and Dahl]{shallue2019measuring}
Shallue, C.~J., Lee, J., Antognini, J., Sohl-Dickstein, J., Frostig, R., and Dahl, G.~E.
\newblock Measuring the effects of data parallelism on neural network training.
\newblock \emph{Journal of Machine Learning Research}, 20\penalty0 (112):\penalty0 1--49, 2019.

\bibitem[Smith \& Le(2017)Smith and Le]{smith2017bayesian}
Smith, S.~L. and Le, Q.~V.
\newblock A bayesian perspective on generalization and stochastic gradient descent.
\newblock \emph{arXiv preprint arXiv:1710.06451}, 2017.

\bibitem[Smith et~al.(2017)Smith, Kindermans, Ying, and Le]{smith2017don}
Smith, S.~L., Kindermans, P.-J., Ying, C., and Le, Q.~V.
\newblock Don't decay the learning rate, increase the batch size.
\newblock \emph{arXiv preprint arXiv:1711.00489}, 2017.

\bibitem[Stephan et~al.(2017)Stephan, Hoffman, Blei, et~al.]{stephan2017stochastic}
Stephan, M., Hoffman, M.~D., Blei, D.~M., et~al.
\newblock Stochastic gradient descent as approximate bayesian inference.
\newblock \emph{Journal of Machine Learning Research}, 18\penalty0 (134):\penalty0 1--35, 2017.

\bibitem[Touvron et~al.(2023)Touvron, Martin, Stone, Albert, Almahairi, Babaei, Bashlykov, Batra, Bhargava, Bhosale, et~al.]{llama}
Touvron, H., Martin, L., Stone, K., Albert, P., Almahairi, A., Babaei, Y., Bashlykov, N., Batra, S., Bhargava, P., Bhosale, S., et~al.
\newblock Llama 2: Open foundation and fine-tuned chat models.
\newblock \emph{arXiv preprint arXiv:2307.09288}, 2023.

\bibitem[Wortsman et~al.(2023)Wortsman, Liu, Xiao, Everett, Alemi, Adlam, Co-Reyes, Gur, Kumar, Novak, et~al.]{wortsman2023small}
Wortsman, M., Liu, P.~J., Xiao, L., Everett, K., Alemi, A., Adlam, B., Co-Reyes, J.~D., Gur, I., Kumar, A., Novak, R., et~al.
\newblock Small-scale proxies for large-scale transformer training instabilities.
\newblock \emph{arXiv preprint arXiv:2309.14322}, 2023.

\bibitem[Zhang et~al.(2019)Zhang, Li, Nado, Martens, Sachdeva, Dahl, Shallue, and Grosse]{zhang2019algorithmic}
Zhang, G., Li, L., Nado, Z., Martens, J., Sachdeva, S., Dahl, G., Shallue, C., and Grosse, R.~B.
\newblock Which algorithmic choices matter at which batch sizes? insights from a noisy quadratic model.
\newblock \emph{Advances in neural information processing systems}, 32, 2019.

\bibitem[Zhang et~al.(2022)Zhang, Roller, Goyal, Artetxe, Chen, Chen, Dewan, Diab, Li, Lin, et~al.]{opt}
Zhang, S., Roller, S., Goyal, N., Artetxe, M., Chen, M., Chen, S., Dewan, C., Diab, M., Li, X., Lin, X.~V., et~al.
\newblock Opt: Open pre-trained transformer language models.
\newblock \emph{arXiv preprint arXiv:2205.01068}, 2022.

\bibitem[Zhang et~al.(2023)Zhang, Han, Cao, Dai, Miao, Cao, Yang, and Xu]{zhang2023adam}
Zhang, Y., Han, Y., Cao, S., Dai, G., Miao, Y., Cao, T., Yang, F., and Xu, N.
\newblock Adam accumulation to reduce memory footprints of both activations and gradients for large-scale dnn training.
\newblock \emph{arXiv preprint arXiv:2305.19982}, 2023.

\end{thebibliography}

\newpage
\onecolumn
\appendix

\section{Connecting limits in Theorem 1 with gradients}
\label{app:grads}

Notice that the quantities in Eq.~\eqref{eq:thm} can be understood as derivatives,
\begin{subequations}
\label{eq:thm:grads}
\begin{align}
  \dl{m_t} &= \lim_{\delta \rightarrow 0} \frac{m_t - m_{t-\kappa}}{\delta} & 
  \dl{m'_t} &= \lim_{\delta \rightarrow 0} \frac{m'_t - m'_{t-\kappa}}{\delta},\\
  \dl{v_t} &= \lim_{\delta \rightarrow 0} \frac{v_t - v_{t-\kappa}}{\delta}&
  \dl{v'_t} &= \lim_{\delta \rightarrow 0} \frac{v'_t - v'_{t-\kappa}}{\delta},\\
  \dl{w_t} &= \lim_{\delta \rightarrow 0} \frac{w_t - w_{t-\kappa}}{\delta} &
  \dl{w'_t} &= \lim_{\delta \rightarrow 0} \frac{w'_t - w'_{t-\kappa}}{\delta}.
\end{align}
\end{subequations}
To obtain this result, we used the observation that for $\delta = 0$, the state does not change, so 
\begin{subequations}
\begin{align}
  w_t(\delta=0)  &= w_{t-\kappa}&
  w'_t(\delta=0) &= w_{t-\kappa},\\
  m_t(\delta=0)  &= m_{t-\kappa}&
  m'_t(\delta=0) &= m_{t-\kappa},\\
  v_t(\delta=0)  &= v_{t-\kappa}&
  v'_t(\delta=0) &= v_{t-\kappa}.
\end{align}
\end{subequations}
Thus, all we need to do is show equivalence of the gradient of the state ($m_t$, $v_t$ and $w_t$) wrt $\delta$.

\section{Warmup by proving a similar result for SGD}

To prove the equivalance of the merged and unmerged optimizers, it is easiest to first consider SGD.
A single SGD step can be written,
\begin{subequations}
\label{eq:sgd_compute_graph}
\begin{align}
  \label{eq:sgd_compute_graph:gt}
  g_{t} &= \grad_t(w_{t-1}) \\
  \label{eq:sgd_compute_graph:wt}
  w_{t} &= \sgd(w_{t-1}, g_{t}, \eta) 
\end{align}
\end{subequations}
where $\grad_t$ is a function that computes the gradients for the $t$th microbatch, and $\sgd$ is a function that applies a single SGD update,
\begin{subequations}
\begin{align}
  \text{grad}_t(w) &= \nabla \mathcal{L}_{t}(w_{t-1}),\\
  \text{sgd}(w, g, \eta) &= (1-\lambda \eta) w - \eta g.
\end{align}
\end{subequations}
We have written the updates in this slightly unusual form to make it possible to carefully distinguish partial and total derivatives, which turns out to be important in our setting to apply the chain rule correctly.
Specifically, the chain rule applied to $g_t$ and $w_t$ gives,
\begin{subequations}
\begin{align}
  \de{g_{t}} &= \dd[g_t]{w_{t-1}} \de{w_{t-1}},\\
  \de{w_{t}} &= \dd[w_t]{w_{t-1}} \de{w_{t-1}} + \dd[w_t]{g_t} \de{g_t} + \dd[w_t]{\eta}.
\end{align}
\end{subequations}
The partial derivatives here are formally defined as,
\begin{subequations}
\begin{align}
\dd[g_t]{w_{t-1}} &= \lim_{h \rightarrow 0} \frac{\grad_t(w_{t-1}+h) - \grad_t(w_{t-1})}{h}\\
\dd[w_t]{w_{t-1}} &= \lim_{h \rightarrow 0} \frac{\sgd(w_{t-1}+h, g_t, \eta) - \sgd(w_{t-1}, g_t, \eta)}{h}\\
\dd[w_t]{g_t} &= \lim_{h \rightarrow 0} \frac{\sgd(w_{t-1}, g_t+h, \eta) - \sgd(w_{t-1}, g_t, \eta)}{h}\\
\dd[w_t]{\eta} &= \lim_{h \rightarrow 0} \frac{\sgd(w_{t-1}, g_t, \eta+h) - \sgd(w_{t-1}, g_t, \eta)}{h}
\end{align}
\end{subequations}
i.e.\ the partial derivatives compute the change in the output of the \textit{function} (either $\grad_t$ or $\sgd$) as one of the arguments to those functions changes, while all the other arguments are held fixed.
In contrast, the total derivatives ($d/d\eta$) represent the total change through the whole ``compute graph'', i.e.\ the total change in $w_t$ and $g_t$ if we start from a fixed $w_{t-\kappa}$ and change $\eta$.

To compute the total derivative for $w_t$, we use the following partial derivatives,
\begin{subequations}
\begin{align}
  \dd[w_t]{w_{t-1}} &= 1-\lambda \eta\\
  \dd[w_t]{g_t} &= - \eta\\
  \dd[w_t]{\eta} &= - g_t - \lambda w_{t-1}.
\end{align}
\end{subequations}
Substituting these partial derivatives into the total derivative,
\begin{align}
  \de{w_{t}} &= (1-\eta \lambda) \de{w_{t-1}} - \eta \de{g_{t}} - g_t - \lambda w_{t-1}.
\end{align}
Evaluating at $\eta=0$,
\begin{align}
  \at{\de{w_{t}}}_{\eta=0} &= \at{\de{w_{t-1}}}_{\eta=0} - g_t - \lambda w_{t-1},
\end{align}
we get a simple recursive expression.

Then, if we fix $w_{t-\kappa}$, and consider computing $w_t$ through $\kappa$ steps of gradient descent, we get,
\begin{align}
  \at{\de{w_{t}}}_{\eta=0} &= - \sum_{k=1}^\kappa \b{g_{t-\kappa + k} + \lambda w_{t - \kappa + k - 1}},
\end{align}
We can simplify this expression by noting that at $\eta = 0$, the weights at all timesteps are equal, $w_{t-1} = w_{t-2} =\dotsm =w_{t-k}$.
This has important implications for the gradient terms.
Specifically, $g_t$ is the gradient for the $t$th datapoint/microbatch, evaluated using the model parameters at timestep $t-1$, i.e.\ $w_{t-1}$.
But if the weights do not change, the average of these gradients can be computed based on the initial set of weights (i.e.\ based on $w_{t-\kappa}$),
\begin{align}
  \label{eq:g'}
  \tfrac{1}{\kappa} \tsum_{k=1}^\kappa g_{t-k+\kappa} \Bigr\rvert_{\eta=0} = 
  \tfrac{1}{\kappa} \tsum_{k=1}^\kappa \grad_{t-k+\kappa}(w_{t-k+\kappa-1}) \Bigr\rvert_{\eta=0}
  &= \tfrac{1}{\kappa }\tsum_{k=1}^\kappa \grad_{t-k+\kappa}(w_{t-\kappa}) = g'_t.
\end{align}
Here, $g'_t$ is this average gradient for data across all $\kappa$ steps, but evaluated using the initial weights, $w_{t-\kappa}$.
Thus the gradient resulting from the $\kappa$ updates is,
\begin{align}
  \label{eq:sgd_final_multistep_grad}
  \at{\de{w_{t}}}_{\eta=0} &= - \kappa \lambda w_{t-\kappa} - \kappa g'_t.
\end{align}
This is starting to resemble a single step of a merged optimizer which performs a single step using a minibatch consisting of the $\kappa$ minibatches.

However, to be sure of the connection to a single step of a merged optimizer, we need to formally define this optimizer.
\begin{subequations}
\begin{align}
  g'_{t} &= \grad'_t(w_{t-\kappa}) \\
  w'_{t} &= \sgd(w_{t-\kappa}, g'_{t}, \eta') 
\end{align}
\end{subequations}
where $\grad'_t$ is a function that computes the gradients for the $t$th \textit{minibatch},
\begin{align}
  \text{grad}'_t(w) &= \tfrac{1}{\kappa} \tsum_{k=1}^\kappa \text{grad}_{t-\kappa + k}(w) 
\end{align}
Now, we compute the gradient of $w'_t$ wrt $\eta$, taking $\eta' = \kappa \eta$,
\begin{align}
  \at{\de{w'_{t}}}_{\eta=0} &= - \kappa \lambda w_{t-\kappa} - \kappa g'_t.
\end{align}
This is exactly equal to the gradient wrt $\eta$ of the multi-step gradient in Eq.~\eqref{eq:sgd_final_multistep_grad}, which establishes our result.

\section{Proof of Theorem~\ref{thm:micro_bi_equiv}}
\label{app:proof_1}
In Appendix~\ref{app:grads}, we established that the limits in Theorem~\ref{thm:micro_bi_equiv} can be understood as derivatives (Eq.~\ref{eq:thm:grads}).

However, computing gradients through multiple Adam updates is not trivial.
To do so correctly, we need to be careful to distinguish partial and total derivatives.
To that end, we write \mc~Adam updates (Alg.~\ref{alg:micro_adamw}) (i.e.\ standard Adam applied to microbatches) in an unusual form,
\begin{subequations}
\label{eq:adam_compute_graph}
\begin{align}
  g_{t} &= \grad_{t}(w_{t-1})\\
  m_{t} &= \emam(m_{t-1}, g_t, \delta)\\
  v_{t} &= \emav(v_{t-1}, g_t, \delta)\\
  u_{t} &= \fu_{t}(m_{t}, v_{t})\\
  w_{t} &= \fw(w_{t-1}, u_t, \delta)
\end{align}
\end{subequations}
where,
\begin{subequations}
\begin{align}
  \emam(m, g, \delta) &= (1-\delta \gb_1) m + \delta \gb_1 g\\
  \emav(v, g, \delta) &= (1-\delta \gb_2) v + \delta \gb_2 g^2\\
  \fw(w, u, \delta) &= (1- \delta \eb \lambda) w - \delta \eb u.\\
  \fu_t(m, v) &= \frac{\mh_t(m)}{\sqrt{\vh_t(v)} + \epsilon}
  \intertext{where,}
  \label{eq:mh}
  \mh_t(m) &= \frac{m}{1-\beta_1^t} = \frac{m}{1-(1-\delta \gb_1)^t}\\
  \label{eq:vh}
  \vh_t(v) &= \frac{v}{1-\beta_2^t} = \frac{v}{1-(1-\delta \gb_2)^t}
\end{align}
\end{subequations}
We are interested in the three state variables, $m_t$, $v_t$ and $w_t$, that propagate across timesteps.
Using the chain rule, the gradients for these variables can be written,
\begin{subequations}
\begin{align}
  \label{eq:adam:dm}
  \dl{m_{t}} &= \dd[m_t]{g_t} \dl{g_t} + \dd[m_t]{m_{t-1}} \dl{m_{t-1}} + \dd[m_t]{\delta}\\
  \label{eq:adam:dv}
  \dl{v_{t}} &= \dd[v_t]{g_t} \dl{g_t} + \dd[v_t]{v_{t-1}} \dl{v_{t-1}} + \dd[v_t]{\delta}\\
  \label{eq:adam:dw}
  \dl{w_{t}} &= \dd[w_t]{w_{t-1}} \dl{w_{t-1}} + \dd[w_t]{u_{t}} \dl{u_{t}} + \dd[w_t]{\delta}
\end{align}
\end{subequations}
We begin by considering the first state variable, $m_t$.
We substitute the partial derivatives,
\begin{subequations}
\begin{align}
  \dd[m_t]{g_t} &= \delta \gb_1\\
  \dd[m_t]{m_{t-1}} &= 1- \delta \gb_1\\
  \dd[m_t]{\delta} &= \gb_1 (g_t - m_{t-1}),
\end{align}
\end{subequations}
into Eq.~\eqref{eq:adam:dm},
\begin{align}
  \dl{m_{t}} &=  \delta \gb_1 \dl{g_t} + (1- \delta \gb_1) \dl{m_{t-1}} + \gb_1 (g_t - m_{t-1}).
\end{align}
Now, we evaluate at $\delta=0$,
\begin{align}
  \at{\dl{m_{t}}}_{\delta=0} &= \at{\dl{m_{t-1}}}_{\delta=0} + \gb_1 (g_t - m_{t-1}),
\end{align}
which gives a simple, recursive result.

We follow almost exactly the same procedure for $v_t$.
We substitute the partial derivatives,
\begin{subequations}
\begin{align}
  \dd[v_t]{g_t} &= 2 \delta \gb_2 g_t,\\
  \dd[v_t]{v_{t-1}} &= 1- \delta \gb_2,\\
  \dd[v_t]{\delta} &= \gb_2 (g_t^2 - v_{t-1}).
\end{align}
\end{subequations}
into Eq.~\eqref{eq:adam:dv},
\begin{align}
  \dl{v_{t}} &= 2 \delta \gb_2 g_t \dl{g_t} + (1- \delta \gb_2) \dl{v_{t-1}} + \gb_2 (g_t^2 - v_{t-1}).
\end{align}
Now, we evaluate at $\delta=0$,
\begin{align}
  \at{\dl{v_{t}}}_{\delta=0} &= \at{\dl{v_{t-1}}}_{\delta=0} + \gb_2 (g_t^2 - v_{t-1}),
\end{align}
which gives a simple, recursive result.

Finally, we follow a similar process for $w_t$. We substitute the partial derivatives,
\begin{subequations}
\begin{align}
  \dd[w_t]{u_t} &= \delta \eb\\
  \dd[w_t]{w_{t-1}} &= 1- \delta \eb \lambda\\
  \dd[w_t]{\delta} &= -\eb (u_t + \lambda w_{t-1}).
\end{align}
\end{subequations}
into Eq.~\eqref{eq:adam:dw},
\begin{align}
  \dl{w_{t}} &= \delta \eb \dl{u_t} + (1-\delta \eb \lambda) \dl{w_{t-1}} - \eb (u_t + \lambda w_{t-1}).
\end{align}
Now, we evaluate at $\delta=0$,
\begin{align}
  \at{\dl{w_{t}}}_{\delta=0} &= \at{\dl{w_{t-1}}}_{\delta=0} - \eb (u_t + \lambda w_{t-1}).
\end{align}
which gives a simple, recursive result.

Now, we consider fixing $m_{t-\kappa}$, $v_{t-\kappa}$, and $w_{t-\kappa}$ and performing $\kappa$ steps of the optimizer
\begin{subequations}
\label{eq:adam_multi_final_grads}
\begin{align}
  \at{\dl{m_{t}}}_{\delta=0} &= \gb_1 \sum_{k=1}^\kappa \b{g_{t-\kappa+k} - m_{t-\kappa+k-1}}\\
  \at{\dl{v_{t}}}_{\delta=0} &= \gb_2 \sum_{k=1}^\kappa \b{g_{t-\kappa+k}^2 - v_{t-\kappa+k-1}}\\
  \at{\dl{w_{t}}}_{\delta=0} &= - \eb \sum_{k=1}^\kappa \b{u_{t-\kappa+k} + \lambda w_{t-\kappa+k-1}}.
\end{align}
\end{subequations}
Remember that as $\delta \rightarrow 0$ all the state variables do not change, and are equal to their initial values,
\begin{subequations}
\begin{align}
  m_{t-\kappa} &= m_{t-\kappa + 1} = \dotsm = m_{t-1} = m_{t}\\
  \spacev_{t-\kappa} &= \spacev_{t-\kappa+1} = \dotsm = \spacev_{t-1} = \spacev_{t}\\
  \spacew_{t-\kappa} &= \spacew_{t-\kappa+1} = \dotsm = \spacew_{t-1} = \spacew_{t}.
\end{align}
\end{subequations}
Using these constant values, along with Eq.~\eqref{eq:g'},
\begin{subequations}
\label{eq:micro_adam_grads_inter}
\begin{align}
  \at{\dl{m_{t}}}_{\delta=0} &= - \kappa \gb_1 m_{t-\kappa} + \gb_1 \tsum_{k=1}^\kappa g_{t-\kappa+k}\\
  \at{\dl{v_{t}}}_{\delta=0} &= - \kappa \gb_2 v_{t-\kappa} + \gb_2 \tsum_{k=1}^\kappa g_{t-\kappa+k}^2\\
  \at{\dl{w_{t}}}_{\delta=0} &= - \eb (\tsum_{k=1}^\kappa u_{t-\kappa + k} + \kappa \lambda w_{t-\kappa}).
\end{align}
\end{subequations}
Finally, we consider the $u_{t-\kappa+k}$ terms, which requires us to consider the debiasing steps.
Debiasing actually stops doing anything for fixed $t$, in the limit as $\delta\rightarrow 0$, (i.e.\ $\mh_{t}(m) =m$).
To avoid this trivial equivalence, we therefore consider a limit in which $t$ increases as $\delta$ falls,
\begin{align}
  t = \tb / \delta,
\end{align}
where $\tb$ is fixed.
That makes sense, because $\delta$ decreases, all the updates become smaller, so we may want to perform more iterations.
We are going to use the standard limit,
\begin{align}
  \lim_{x \rightarrow \infty} (1+\tfrac{a}{x})^{bx} &= e^{ab}.
\end{align}
In particular, considering the $(1-\gb_1 \delta)^t$ term in Eq.~\eqref{eq:mh}, taking $t=\tb/\delta$, $\delta = 1/x$, and identifying $a=-\kappa \gb$ and $b=\tb$,
\begin{align}
\lim_{\delta \rightarrow 0} (1- \gb_1 \delta )^{t} &= 
\lim_{x \rightarrow \infty} (1-\tfrac{\gb_1}{x})^{\tb x} = e^{-\gb_1 \tb}.\\
\lim_{\delta \rightarrow 0} (1- \gb_2 \delta )^{t} &= 
\lim_{x \rightarrow \infty} (1-\tfrac{\gb_2}{x})^{\tb x} = e^{-\gb_2 \tb}.
\end{align}
(where the second equation is exactly the same thing, but for $v$).
In this limit, the debiasing steps can be written,
\begin{align}
  \lim_{\delta \rightarrow 0} \mh_{t}(m) &= e^{-\gb_1 \tb} m,\\
  \lim_{\delta \rightarrow 0} \vh_{t}(v) &= e^{-\gb_2 \tb} v.
\end{align}
Substituting the result of these debiasing steps into Eq.~\eqref{eq:micro_adam_grads_inter},
\begin{subequations}
\begin{align}
  \at{\dl{m_{t}}}_{\delta=0} &= - \kappa \gb_1 m_{t-\kappa} + \gb_1 \tsum_{k=1}^\kappa g_{t-\kappa+k}\\
  \at{\dl{v_{t}}}_{\delta=0} &= - \kappa \gb_2 v_{t-\kappa} + \gb_2 \tsum_{k=1}^\kappa g_{t-\kappa+k}^2\\
  \at{\dl{w_{t}}}_{\delta=0} &= - \kappa \eb \b{\frac{u_{t-\kappa} e^{-\gb_1 \tb }}{\sqrt{v_{t-\kappa}e^{-\gb_2 \tb}} + \epsilon} + \lambda w_{t-\kappa}}.
\end{align}
\end{subequations}
Overall, these updates resembles a single step of the merged, batch size invariant Adam.

But to be sure, we write batch size invariant Adam updates in Alg.~\eqref{alg:invariant_adamw} as,
\begin{subequations}
\begin{align}
  g'_{t} &= \grad'_t(w_{t-\kappa})\\
  s'_{t} &= \gradsq'_t(w_{t-\kappa})\\
  m_t' &= \emam'(m_{t-\kappa}, g'_t, \delta)\\
  v_t' &= \emav'(v_{t-\kappa}, s'_t, \delta)\\
  u_t' &= \text{norm\_update}'_{t/\kappa}(m'_{t}, v'_{t})\\
  w_t' &= \fw'(w'_{t-1}, u'_t, \delta)
\end{align}
\end{subequations}
where,
\begin{subequations}
\begin{align}
  \grad'_{t}(w_{t-\kappa}) &= \tfrac{1}{\kappa} \tsum_{k=1}^\kappa \grad_{t-k+\kappa}(w_{t-\kappa}),\\
  \gradsq'_{t}(w_{t-\kappa}) &= \tfrac{1}{\kappa} \tsum_{k=1}^\kappa \grad^2_{t-k+\kappa}(w_{t-\kappa}),\\
  \emam'(m, g, \delta) &= (1-\delta \kappa \gb_1) m + \delta \kappa \gb_1 g\\
  \emav'(\spacev, s, \delta) &= (1-\delta \kappa \gb_2) \spacev + \delta \kappa \gb_2 s\\
  \fw'(w, u, \delta) &= (1- \delta \kappa \eb \lambda) w - \delta \kappa \eb u.\\
  \text{norm\_update}'_{t/\kappa}(m, v) &= \frac{\mh'_{t/\kappa}(m)}{\sqrt{\vh'_{t/\kappa}(v)} + \epsilon}
  \intertext{where,}
  \mh'_{t/\kappa}(m) &= \frac{m}{1-(1-\gamma_1')^{t/\kappa}}\\
  \vh'_{t/\kappa}(v) &= \frac{v}{1-(1-\gamma_2')^{t/\kappa}}
\end{align}
\end{subequations}
Now, the gradient of the state variables for the merged, batch size invariant Adam updates is,
\begin{subequations}
\begin{align}
  \at{\dl{m_{t}'}}_{\delta=0} &= \kappa \gb_1 m_{t-\kappa} + \gb_1 \tsum_{k=1}^\kappa g_{t-k + \kappa}\\
  \at{\dl{v_{t}'}}_{\delta=0} &= \kappa \gb_2 v_{t-\kappa} + \gb_2 \tsum_{k=1}^\kappa g_{t-k + \kappa}^2\\
  \at{\dl{w_{t}'}}_{\delta=0} &= - \kappa \eb (u_{t} + \lambda w_{t-\kappa}).
\end{align}
\end{subequations}
Comparing these expressions to Eq.~\eqref{eq:adam_multi_final_grads}, we are almost done.
It only remains to consider $\kappa u_t$, again in a limit in which $t$ increases as $\delta$ decreases, $t = \tb / \delta$.
Again, we consider the $(1-\gamma_1')^{t/\kappa}$ term, using the standard limit, $\lim_{x \rightarrow \infty} (1+\tfrac{a}{x})^{bx} = e^{ab}$, take $\delta = 1/x$, and identify $a=-\gb$ and $b=\tb$,
\begin{align}
\lim_{\delta \rightarrow 0} (1-\gamma_1')^{\tb/(\delta \kappa)} &= \lim_{\delta \rightarrow 0} (1-\kappa \gb_1 \delta )^{\tb/(\delta \kappa)} = 
\lim_{x \rightarrow \infty} (1-\tfrac{\kappa \gb_1}{x} )^{\tfrac{\tb}{\kappa}x} = e^{-\gb_1 \tb}.
\end{align}
Thus, the debiasing steps can be written,
\begin{subequations}
\begin{align}
  \lim_{\delta \rightarrow 0} \mh'_{t/\kappa}(m) &= e^{-\gb_1 \tb} m,\\
  \lim_{\delta \rightarrow 0} \vh'_{t/\kappa}(v) &= e^{-\gb_2 \tb} v.
\end{align}
\end{subequations}
Which gives final gradients,
\begin{subequations}
\begin{align}
  \at{\dl{m'_{t}}}_{\delta=0} &= - \kappa \gb_1 m_{t-\kappa} + \gb_1 \tsum_{k=1}^\kappa g_{t-\kappa+k}\\
  \at{\dl{v'_{t}}}_{\delta=0} &= - \kappa \gb_2 v_{t-\kappa} + \gb_2 \tsum_{k=1}^\kappa g_{t-\kappa+k}^2\\
  \at{\dl{w'_{t}}}_{\delta=0} &= - \kappa \eb \b{\frac{u_{t-\kappa} e^{-\gb_1 \tb }}{\sqrt{v_{t-\kappa}e^{-\gb_2 \tb}} + \epsilon} + \lambda w_{t-\kappa}}.
\end{align}
\end{subequations}
These gradients are exactly the same as the gradients for the multi-step updates (Eq.~\ref{eq:adam_multi_final_grads}), which proves our result.

\section{Proof of Theorem~\ref{thm:batch size_invariance}}
\label{app:proof_2}

By Theorem 1,
\begin{subequations}
\begin{align}
  \lim_{\delta \rightarrow 0} \frac{w_t' - w_{t-K}}{\delta} &= 
  \lim_{\delta \rightarrow 0} \frac{w_t - w_{t-K}}{\delta} =
  \lim_{\delta \rightarrow 0} \frac{w_t'' - w_{t-K}}{\delta}\\
  \lim_{\delta \rightarrow 0} \frac{m_t'- m_{t-K}}{\delta} &= 
  \lim_{\delta \rightarrow 0} \frac{m_t- m_{t-K}}{\delta} = 
  \lim_{\delta \rightarrow 0} \frac{m_t'' - m_{t-K}}{\delta}\\
  \lim_{\delta \rightarrow 0} \frac{v_t' - v_{t-K}}{\delta} &= 
  \lim_{\delta \rightarrow 0} \frac{v_t - v_{t-K}}{\delta} = 
  \lim_{\delta \rightarrow 0} \frac{v_t'' - v_{t-K}}{\delta}.
\end{align}
\end{subequations}
where $m_t, v_t, w_t$ is the result of applying $K$ steps of Micro-Adam, with hyperparameters $\eta, \gamma_1, \gamma_2$.

\end{document}